%% file: iclr2026_conference.tex
\documentclass{article} % For LaTeX2e
\usepackage{iclr2026_conference,times}

\input{math_commands.tex}

\usepackage{hyperref}
\usepackage{url}
\usepackage{booktabs}
\usepackage{multirow}
\usepackage{graphicx}
\iclrfinalcopy

\title{Guidance Is Not a Hyperparameter: \\Learning Dynamic Control in Diffusion Language Models}

\author{Fan Zhou, Tim Van de Cruys\\
KU Leuven\\
\texttt{\{fan.zhou,tim.vandecruys\}@kuleuven.be} \\
}

\begin{document}

\maketitle

\begin{abstract}
Classifier-Free Guidance (CFG) is a widely used mechanism for controlling diffusion-based generative models, yet its guidance scale is typically treated as a fixed hyperparameter throughout generation. This static design yields a suboptimal controllability–quality tradeoff, as the optimal degree of guidance varies across tasks and across different stages of the diffusion process, especially in NLP domain. We recast CFG scale selection as a sequential decision-making problem and propose to learn dynamic guidance trajectories via reinforcement learning. Specifically, we model the guidance scale as a discrete control action selected at each generation step based on the evolving diffusion state, and optimize a policy using Proximal Policy Optimization (PPO) under task-level rewards. Experiments on three controlled NLP generation tasks using discrete diffusion language models demonstrate that adaptive guidance consistently achieves a better balance between controllability and generation quality than fixed-scale strategies. Further analysis of the learned policies reveals distinct and interpretable guidance trajectories across tasks, underscoring the importance of treating guidance as a dynamic control process rather than a static design choice.

\end{abstract}

\section{Introduction}

Diffusion-based generative models have emerged as a dominant paradigm across multiple domains. In computer vision, they have become the standard approach for high-fidelity image generation \citep{rombach2022high}, while recent advances in NLP domain, especially the discrete diffusion language models (dLLMs) \citep{nie2025large, ye2025dream} have demonstrated their effectiveness for text generation, providing a viable alternative to conventional autoregressive language models. A key factor underlying this success is the ability to explicitly control the generation process through guidance mechanisms. Among these, classifier-free guidance (CFG) \citep{ho2022classifier} has become the de facto standard, exposing a single control variable—the guidance scale—that directly modulates the influence of conditioning information during generation. By adjusting this scale, CFG enables controllable generation but also induces an inherent tradeoff between adherence to the desired conditions and generation quality, suggesting that effective control hinges on how this tradeoff is managed throughout the generation process.

While classifier-free guidance provides a simple and effective control mechanism, its guidance scale is almost always treated as a fixed hyperparameter throughout the entire generation process \citep{ho2022classifier, dhariwal2021diffusion}. In practice, this scale is typically selected via offline tuning and then applied uniformly across all generation steps and inputs \citep{nichol2021improved}. However, diffusion-based generation is inherently sequential and context-dependent: the role and effect of guidance vary across generation stages, and the balance between controllability and generation quality can differ substantially across tasks and inputs\citep{wang2024analysis, zhang2025much}. As a result, applying a single, globally fixed guidance scale introduces a structural mismatch between the stage-wise requirements of the generation process and the control mechanism, leading to suboptimal controllability–quality tradeoffs in practice.

To address the limitations of static guidance, prior work—particularly in computer vision—has explored heuristic strategies for adjusting guidance strength during diffusion sampling, typically by prescribing a fixed functional schedule shared across generation scenarios, such as exponential curves \citep{gaobeyond} or distributions parameterized by simple forms (e.g., Beta schedules) \citep{malarz2025classifier}. While these heuristics can be effective in specific settings, they fundamentally rely on pre-defined schedules that are determined prior to sampling and remain independent of the evolving generation state or task-level objectives. As a result, they implicitly assume that a single schedule—or a small family of hand-crafted schedules—can generalize across diverse inputs, tasks, and controllability requirements. In practice, however, optimal guidance behavior often varies substantially across scenarios, making it difficult for static heuristic schedules to capture task-specific needs without extensive manual tuning, particularly in NLP tasks with heterogeneous objectives. More importantly, such heuristic approaches lack a systematic mechanism for incorporating task-level feedback during generation, highlighting the need for a more flexible and adaptive approach to guidance control.

These observations suggest that guidance control in diffusion models is inherently a sequential decision-making problem. At each generation step, the choice of guidance strength influences not only the current output but also the future trajectory of the generation process, with its effect only fully reflected in task-level outcomes at the end of sampling. Importantly, there is no oracle supervision for selecting guidance strength at intermediate steps, and downstream task objectives are typically non-differentiable and only observable upon completion of the generation process. As a result, manually specifying guidance schedules or relying on myopic, step-wise rules is insufficient for optimizing long-term task performance. Reinforcement learning provides a principled framework for this setting by enabling sequential decisions to be optimized under sparse and delayed rewards while accounting for long-term effects across generation steps. In this work, we adopt a policy-based reinforcement learning approach to adaptively select guidance scales during generation, enabling dynamic control that responds to the evolving generation context.

In this work, we focus on learning task-specific dynamic guidance trajectories for diffusion-based text generation, rather than adapting guidance on a per-instance basis, which would require reliable instance-level feedback during generation and risk overfitting to noisy trajectory outcomes. Our goal is to capture characteristic guidance behaviors that are optimal for a given generation task and can be consistently applied across different inputs. We instantiate the proposed reinforcement learning formulation in discrete diffusion language models, using LLaDA \citep{nie2025large} as the underlying generative model. We study three representative NLP scenarios with distinct controllability objectives: keyword-conditioned sentence generation, sentiment-controlled style transfer, and length-controlled sentence rewriting. For each task, we optimize a guidance policy using task-specific rewards and analyze the resulting guidance trajectories to understand how controllability–quality tradeoffs evolve across diffusion steps.

Beyond overall generation quality, our primary goal is to understand how guidance behaviors evolve over diffusion steps to effectively balance controllability and generation quality. By learning task-specific guidance policies, we aim to characterize the structure of guidance trajectories that emerge under different controllability objectives. In particular, we seek to answer the following questions: how optimal guidance behaviors differ across tasks, how guidance strength evolves over the course of generation, and to what extent learned guidance trajectories provide insights beyond fixed or heuristic schedules.

Taken together, these considerations motivate our formulation of adaptive guidance as a learning problem. In the following sections, we detail our reinforcement learning approach and present empirical results that illustrate the resulting task-specific guidance behaviors across different controllability objectives.

\section{Related work}

\subsection{Controllability–quality tradeoffs in generation}
A fundamental challenge in controllable text generation is balancing controllability with fluency. 
Prior work has repeatedly observed that enforcing stronger control over attributes such as sentiment, keywords, or style often degrades fluency and naturalness, leading to repetition or semantic drift \citep{hu2017toward, dathathri2019plug, krause2021gedi}. 
This tradeoff has been studied across different generation paradigms. 
In autoregressive language models, recent work primarily relies on generation-time control mechanisms such as weighted decoding, critic-guided decoding, prefix-based control, and instruction-based prompting \citep{zhou2023controlled, kim2023critic, pei2023preadd, zhang2023semantic, shin2025eco}. 
In diffusion-based text generation, controllability is commonly introduced through guidance mechanisms that modify sampling dynamics, including plug-and-play approaches based on external classifiers \citep{horvitz2024paraguide}, classifier-free guidance (CFG) via scaling conditional signals during sampling \citep{ho2022classifier}, as well as training-time methods that incorporate control signals during model learning \citep{zhou2025controllable}. In contrast to these approaches, our work focuses on generation-time controllability in diffusion-based language models and explicitly studies how guidance strength should be allocated over the diffusion process to balance controllability and generation quality, rather than modifying model parameters or decoding objectives.

\subsection{CFG guidance scheduling for control}
Recent work revisits classifier-free guidance (CFG) by treating the guidance scale as a dynamic control variable rather than a fixed hyperparameter. A common approach is to design time-dependent heuristic schedulers that vary guidance strength across diffusion steps, often supported by empirical studies on “how much to guide” and by analyses that explain why different noise regimes favor different guidance magnitudes \citep{wang2024analysis, zhang2025much, malarz2025classifier}. Beyond purely step-based schedules, recent methods incorporate online feedback signals (e.g., attribute confidence or constraint satisfaction) to adapt guidance strength on-the-fly during sampling, providing sample-level adjustment without learning a controller \citep{papalampidi2025dynamic}. In parallel, theory- or control-inspired formulations revisit CFG in discrete diffusion settings and derive more principled adaptive scaling rules from modeling assumptions or hand-designed objectives, including theory-informed improvements for discrete diffusion and stochastic optimal control perspectives \citep{rojas2025theory, azangulov2025adaptive}. While these methods introduce adaptive or scheduled guidance during diffusion sampling, they primarily rely on hand-designed rules, instantaneous feedback signals, or analytical scaling laws, and do not explicitly learn a guidance policy optimized for task-level objectives. In contrast, our approach formulates guidance selection as a sequential decision-making problem and learns task-specific guidance trajectories via reinforcement learning under sparse, terminal rewards.

 % \subsection{Reinforcement learning for diffusion models}         Recent work has applied reinforcement learning to optimize
 %  diffusion models for downstream objectives.                   
 %  DDPO~\citep{black2024training} formulates the denoising
 %  process as a multi-step MDP and uses policy gradient methods   to fine-tune diffusion model parameters toward human           preference rewards.
 %  DPOK~\citep{fan2024reinforcement} extends this with
 %  KL-regularized policy optimization to prevent reward
 %  overoptimization.
 %  These methods and subsequent variants~\citep{clark2023directly,
 %  xu2024imagereward} have been primarily developed for
 %  image generation and operate by updating the diffusion model
 %  weights themselves.
 %  In contrast, our approach keeps the pretrained diffusion
 %  language model entirely frozen and learns only a lightweight
 %  external controller that modulates the guidance scale at each
 %  sampling step.
 %  This decoupled design avoids catastrophic forgetting and
 %  expensive model fine-tuning, and enables the same base model
 %  to serve different tasks by simply swapping the guidance policy.

\section{Preliminaries}
\subsection{Discrete Diffusion Language Models}

We consider a discrete diffusion language model (ddLM) defined over token sequences ~\citep{lou2023discrete, nie2025large}.
Let $x_0 = (x_0^1, \dots, x_0^L)$ denote a clean sequence of length $L$, and let $M$ be a special
\textsc{[MASK]} token.
ddLM defines a continuous-time forward noising process that independently masks each token.

Given a noise level $t \in [0,1]$, the forward process is defined as
\begin{equation}
q_t(x_t^i \mid x_0^i) =
\begin{cases}
t, & x_t^i = M, \\
1 - t, & x_t^i = x_0^i ,
\end{cases}
\qquad
q_t(x_t \mid x_0) = \prod_{i=1}^L q_t(x_t^i \mid x_0^i),
\label{eq:forward_mask}
\end{equation}
where $x_t$ denotes the corrupted sequence at noise level $t$.

The reverse process is parameterized by a neural \emph{mask predictor}
$p_\theta(\cdot \mid x_t)$, which models the conditional distribution of the original token at each
masked position given the partially masked sequence.

For conditional generation, ddLM models the distribution $p_\theta(r_0 \mid p_0)$ by applying the
forward masking process only to the response segment while keeping the prompt fixed.
At inference time, generation is performed by simulating the reverse diffusion process from a fully
masked response.
Let $1 = t_K > t_{K-1} > \cdots > t_0 = 0$ denote a discretization of the time interval.
At each reverse step, the mask predictor fills in the masked tokens, followed by a remasking operation
that ensures consistency with the forward process, yielding a valid reverse diffusion process.

 \subsection{Classifier-free Guidance}

Classifier-free guidance (CFG) ~\citep{ho2022classifier} controls conditional diffusion sampling by amplifying
the effect of conditioning information during the reverse process.
Let $c$ denote the condition (e.g., a prompt).
For a corrupted sequence $x_t$, the mask predictor produces token-level log-probabilities
\begin{equation}
\ell_{\text{cond}}^i(x)=\log p_\theta(x \mid x_t, c),
\qquad
\ell_{\text{uncond}}^i(x)=\log p_\theta(x \mid x_t),
\end{equation}
for each masked position $i$.

In dLLMs, CFG is implemented by defining guided logits as
\begin{equation}
\ell_{\text{CFG}}^i(x)
=
\ell_{\text{uncond}}^i(x)
+
(1+\gamma)\big(
\ell_{\text{cond}}^i(x)
-
\ell_{\text{uncond}}^i(x)
\big),
\label{eq:cfg_dllm}
\end{equation}
where $\gamma \ge 0$ is the guidance scale.

The guidance scale $\gamma$ is an inference-time control variable that can be adjusted
during sampling to modify the reverse diffusion dynamics without changing the learned model parameters.

\subsection{Policy Learning for Sequential Decision}
\label{sec:prelim_rl}

We formulate guidance selection as a policy-based sequential decision-making problem,
where a parameterized policy $\pi_\phi(a \mid s)$ selects actions based on the current state
and induces a trajectory of decisions.
The objective is to maximize the expected cumulative reward over a trajectory.

In this work, we optimize the policy using Proximal Policy Optimization (PPO)~\citep{schulman2017proximal},
a standard on-policy algorithm for discrete action spaces.

Policy-based methods such as PPO are well suited for sequential decision problems
with stochastic dynamics and delayed rewards, which motivates their use in our setting.

\section{Method}
\subsection{Motivation: Why Reinforcement Learning}

We start from the observation that different downstream tasks evaluate generated outputs using
distinct criteria.
As a result, we assume that the optimal guidance schedule over diffusion steps is task-dependent,
and no single guidance curve is universally optimal across tasks.

Importantly, our goal is not to fit guidance decisions to individual samples.
Instead, we aim to learn guidance strategies that generalize across samples drawn from the same task
distribution.
This task-level perspective is necessary to avoid overfitting to noisy instance-level outcomes and
to enable robust generalization.
Formally, let $\mathcal{T}$ denote a distribution over tasks, and let $\tau$ represent a complete
diffusion sampling trajectory induced by a guidance strategy.
Our objective is to maximize the expected task-level reward:
\begin{equation}
\max_{\pi} \;
\mathbb{E}_{\mathcal{T}}
\left[
\mathbb{E}_{\tau \sim \pi(\mathcal{T})}
\left[
R(\tau)
\right]
\right],
\label{eq:task_objective}
\end{equation}
where $R(\tau)$ evaluates the quality of the final generated output.

Many existing guidance strategies are motivated by heuristic intuitions, such as using smaller
guidance scales at highly noisy stages, larger scales at intermediate steps, and smaller scales
again near convergence.
Such intuitions often lead to parameterized guidance curves with a small number of degrees of
freedom.
Let a guidance curve be parameterized as $g_\theta(t)$ with parameters $\theta \in \mathbb{R}^d$.
Searching for a task-specific guidance strategy then amounts to solving the following optimization
problem:
\begin{equation}
\theta^\star
=
\arg\max_{\theta}
\;
\mathbb{E}_{\tau \sim g_\theta}
\left[
R(\tau)
\right].
\label{eq:curve_search}
\end{equation}

However, extending these approaches to task-specific guidance quickly becomes computationally
infeasible.
Searching over flexible curve families significantly enlarges the action space, and evaluating each
candidate curve requires multiple diffusion sampling runs.
Given the high computational cost and low sampling efficiency of discrete diffusion language models,
curve-level search in Eq.~\eqref{eq:curve_search} is prohibitively expensive in practice.

Purely heuristic guidance strategies further rely on fixed, deterministic schedules of the form
$a_k = g(k)$ and do not incorporate task-level feedback.
Moreover, there is no oracle supervision for optimal guidance decisions at intermediate diffusion
steps, and task evaluation is only available upon completion of the sampling process.
As a result, selecting guidance actions based solely on intuition or static rules is insufficient
for task-adaptive control.

Taken together, these considerations indicate that adaptive guidance selection should be treated as
a sequential decision-making problem with delayed and noisy feedback.
This naturally motivates a reinforcement learning formulation that directly optimizes expected
trajectory-level returns.

\subsection{Policy Learning with PPO}

Following the formulation in Section~4.1, we model adaptive guidance selection during diffusion
sampling as a policy learning problem.
A guidance policy interacts with the diffusion process over multiple steps, producing a sequence
of guidance decisions that jointly determine the final generated output.

\paragraph{Policy Formulation.}
Let $\tau = (s_K, a_K, \dots, s_1, a_1, s_0)$ denote a complete diffusion sampling trajectory, where
$s_k$ represents the sampling state at diffusion step $k$ and $a_k$ denotes the guidance action
selected at that step.
A policy $\pi_\phi(a_k \mid s_k)$ maps the current sampling state to a guidance action.
One complete diffusion run corresponds to a single episode.

The objective of policy learning is to maximize the expected cumulative reward over trajectories:
\begin{equation}
J(\pi_\phi)
=
\mathbb{E}_{\tau \sim \pi_\phi}
\left[
\sum_{k=0}^{K} r_k
\right],
\label{eq:policy_objective}
\end{equation}
where $r_k$ denotes the reward received at diffusion step $k$.

\paragraph{Sparse Reward Design.}
We adopt a sparse reward formulation aligned with downstream task evaluation.
Specifically, rewards are assigned only at the final diffusion step:
\begin{equation}
r_k =
\begin{cases}
R(\tau), & k = 0, \\
0, & k > 0,
\end{cases}
\label{eq:sparse_reward}
\end{equation}
where $R(\tau)$ evaluates the quality of the completed generated sequence.

Downstream tasks define their objectives solely on the final output.
In contrast, intermediate diffusion states $s_k$ contain masked tokens and stochastic noise and do
not correspond to semantically interpretable text.
As a result, assigning step-wise rewards based on such states requires evaluating the final task
objective using partial and noisy representations.

Formally, for any step-wise reward $\hat{r}_k = f(s_k)$ derived from an intermediate state, the best
possible predictor of the final task reward $R(\tau)$ in the mean-squared sense is the conditional
expectation $\mathbb{E}[R(\tau)\mid s_k]$.
The informativeness of any step-wise reward is therefore governed by the variance of this quantity:
\begin{equation}
\mathrm{Var}(R)
=
\mathbb{E}\!\left[\mathrm{Var}(R \mid s_k)\right]
+
\mathrm{Var}\!\left(\mathbb{E}[R \mid s_k]\right),
\label{eq:variance_decomposition}
\end{equation}
where $\mathrm{Var}(\mathbb{E}[R \mid s_k])$ measures the portion of reward variability that can be
explained by the intermediate state $s_k$.

This explainable variance is fundamentally limited by the mutual information between $s_k$ and the
final output.
Since $R(\tau)$ is a deterministic function of the final generated sequence $x_0$, the data
processing inequality yields
\begin{equation}
I(R; s_k) \le I(x_0; s_k),
\label{eq:dpi}
\end{equation}
where $I(\cdot;\cdot)$ denotes mutual information.
At early and intermediate diffusion steps, $s_k$ remains highly noisy and masked, implying limited
mutual information with the final output $x_0$.
Consequently, $\mathrm{Var}(\mathbb{E}[R \mid s_k])$ is small, and any step-wise reward derived from
$s_k$ is inherently noisy and weakly correlated with the true task objective.

For these reasons, dense step-wise rewards are not considered.
Using sparse terminal rewards avoids introducing proxy objectives based on low-information
intermediate states and directly aligns policy optimization with the true task-level evaluation
defined in Eq.~\eqref{eq:task_objective}.

\paragraph{Action Repetition and Temporal Abstraction.}
As the number of diffusion steps $K$ increases, sparse terminal rewards in
Eq.~\eqref{eq:sparse_reward} induce long-horizon credit assignment with high variance.
To reduce the effective decision horizon, we adopt action repetition, where a single guidance
action is held constant for $n$ consecutive diffusion steps.
Let $m=\lceil K/n\rceil$ denote the number of decision blocks, and let $\tilde{a}_j$ be the action
selected for block $j\in\{1,\dots,m\}$.
The per-step action is defined as
\begin{equation}
a_k = \tilde{a}_{\lceil k/n\rceil}.
\label{eq:action_repetition}
\end{equation}

This temporal abstraction reduces the number of policy decisions from $K$ to $m$, mitigating the
variance of policy updates under sparse terminal feedback.
Moreover, holding actions constant over short intervals encourages smooth guidance trajectories by
construction, preventing abrupt changes between consecutive diffusion steps.

\paragraph{Discrete Action Space for Generalized Trajectories.}
Our goal is to learn task-generalized guidance behaviors characterized by coarse trajectory
properties, such as early-, mid-, and late-stage guidance strength, overall trends, and peak values,
rather than fine-grained per-step control.
Accordingly, we restrict the action space to a small discrete set of guidance scales:
\begin{equation}
\mathcal{A}=\{0, 0.25, 0.5, 0.75, 1.0, 1.25, 1.5, 1.75, 2.0, 2.25, 2.5, 2.75, 3.0\},
\qquad |\mathcal{A}|=13.
\label{eq:discrete_actions}
\end{equation}

This discrete design provides sufficient expressive power to capture distinct guidance regimes
across diffusion stages while substantially reducing exploration complexity compared to continuous
control.
In practice, a categorical policy over a finite action set yields more stable on-policy updates
under sparse and noisy trajectory-level rewards.

\paragraph{Policy Optimization with PPO.}

To optimize the guidance policy under sparse and delayed terminal rewards, we adopt Proximal Policy
Optimization (PPO).
We use the standard clipped surrogate objective described in
Section~\ref{sec:prelim_rl}, which provides stable on-policy updates under the stochastic dynamics
of diffusion sampling.
PPO naturally supports discrete action spaces and long-horizon trajectories, making it well suited
for adaptive guidance selection without requiring differentiability of the task-level reward.

\subsection{Sampling Task-Generalized Guidance Trajectories}

After training, the learned guidance policy $\pi_\phi$ does not produce a single deterministic
guidance schedule.
Instead, it induces a distribution over guidance trajectories through stochastic policy sampling
and the inherent randomness of diffusion sampling.
Each rollout corresponds to one realization of guidance decisions over diffusion steps.

\paragraph{Policy-Induced Trajectory Distribution.}
Recall that policy optimization in Eq.~\eqref{eq:task_objective} aims to maximize the expected
task-level reward under the policy-induced trajectory distribution.
Let $\tilde{\tau}^{(i)} = \{\tilde{a}^{(i)}_1,\dots,\tilde{a}^{(i)}_m\}$ denote the sequence of guidance
actions selected for the $m$ decision blocks defined in Eq.~\eqref{eq:action_repetition} during the
$i$-th rollout.
Sampling the learned policy yields a set of trajectories
$\{\tilde{\tau}^{(i)}\}_{i=1}^{N}$ drawn i.i.d.\ from the distribution induced by $\pi_\phi$.

Under this formulation, no single trajectory is expected to be optimal.
Instead, the policy captures a distribution over guidance behaviors that collectively maximize
expected task performance.

\paragraph{Monte Carlo Estimation via Mean Trajectory.}
To obtain a task-generalized guidance strategy, we estimate the expected guidance behavior under the
learned policy by aggregating sampled trajectories.
Specifically, we compute the empirical mean trajectory:
\begin{equation}
\bar{a}_j
=
\frac{1}{N}
\sum_{i=1}^{N}
\tilde{a}^{(i)}_j,
\qquad j=1,\dots,m,
\label{eq:mean_trajectory}
\end{equation}
where $\tilde{a}^{(i)}_j$ denotes the action selected for decision block $j$ in the $i$-th rollout.

Equation~\eqref{eq:mean_trajectory} constitutes a Monte Carlo estimator of the expected guidance
action at each decision block.
Under standard assumptions, the estimator is unbiased and converges to the policy-induced
expectation as the number of sampled trajectories increases.
Averaging across trajectories reduces variance arising from stochastic sampling while preserving
the policy’s task-level generalization.

\paragraph{Frequency-Weighted Monte Carlo Aggregation.}
While uniform averaging treats all sampled trajectories equally, low-frequency trajectories are
more likely to arise from stochastic exploration under sparse terminal rewards.
To emphasize guidance patterns consistently selected by the policy, we further consider a
frequency-weighted aggregation.

Let $f_i$ denote the empirical frequency of the $i$-th guidance interval or pattern across sampled
trajectories, and let $\bar{v}_i$ be the average guidance value associated with that interval.
We define the frequency-weighted mean as
\begin{equation}
\mathrm{CFG}^{\mathrm{freq}}
=
\frac{\sum_{i=1}^{K} f_i^{\,p}\,\bar{v}_i}
{\sum_{i=1}^{K} f_i^{\,p}},
\label{eq:freq_weighted_mean}
\end{equation}
where $p \ge 1$ controls the strength of frequency amplification.

This formulation can be interpreted as a power-transformed Monte Carlo estimator of the expected
guidance behavior under the empirical trajectory distribution.
Using $p>1$ biases aggregation toward dominant guidance patterns while suppressing low-frequency
trajectories that are more likely to reflect exploratory noise.
In practice, this provides a smooth compromise between uniform expectation and mode selection,
yielding robust task-generalized guidance trajectories.

\paragraph{Inference-Time Application.}
The resulting mean or frequency-weighted mean trajectory defines a deterministic, task-generalized
guidance schedule.
This schedule can be applied directly at inference time without additional policy sampling or
optimization, providing a stable approximation of the policy’s expected behavior that remains
fully aligned with the training objective.

\section{Experimental Setup}
\label{sec:exp_setup}

\subsection{Tasks and Evaluation}

We evaluate our method on three controlled text generation tasks that exhibit distinct
controllability--quality trade-offs and require different guidance behaviors across diffusion
stages.
These tasks serve as representative testbeds for assessing task-adaptive guidance strategies,
as no single static guidance schedule is expected to perform optimally across all settings.

\textbf{Keyword-conditioned sentence generation.}
The model is required to generate a fluent sentence that contains all keywords from a given set
of 10 words.
Controllability is measured by the success rate of including all required keywords.
Generation quality is assessed by fluency, measured using perplexity (PPL) computed with a
pretrained GPT-2 language model.

\textbf{Length-controlled generation.}
The model rewrites an input sentence such that the output length falls within 40\%--80\% of the
original sentence length measured in number of words.
Controllability is measured by whether the generated output satisfies the length constraint.
Generation quality is evaluated using content preservation and GPT-2 perplexity.

\textbf{Sentiment style transfer.}
Given an input sentence, the model rewrites it to the opposite sentiment while preserving the
original content.
We consider both positive-to-negative and negative-to-positive transfer.
Controllability is measured by sentiment transfer accuracy using a pretrained binary sentiment
classifier.
Generation quality is assessed by content preservation and fluency, measured by GPT-2
perplexity.

All metrics are computed on the final generated outputs and averaged over the evaluation set.
Training and evaluation are conducted on disjoint prompt sets sampled from the same task
distribution to assess task-level generalization.

\subsection{Reward Definition}

For all tasks, we adopt a sparse terminal reward aligned with downstream task evaluation.
The reward is defined as a weighted combination of controllability and generation quality:
\begin{equation}
R(\tau) = \lambda_1 \, R_{\mathrm{ctrl}}(\tau) + \lambda_2 \, R_{\mathrm{PPL}}(\tau) (+ \lambda_3 \, R_{\mathrm{semantic}}(\tau)),
\end{equation}
where $R_{\mathrm{ctrl}}$ denotes the task-specific controllability score, 
$R_{\mathrm{PPL}}$ denotes the generation fluency and $R_{\mathrm{semantic}}$ denotes the semantic preservation.
Both components are computed only on the final generated output.

\subsection{Baselines}

We compare our method against a set of non-learning guidance baselines that rely on fixed,
predefined guidance schedules and do not adapt guidance decisions based on task-level feedback.

\textbf{Constant guidance}
A fixed classifier-free guidance scale is applied uniformly across all diffusion steps.

\textbf{Heuristic guidance schedules}
We evaluate commonly used hand-crafted guidance curves, including linear increase, linear
decrease, cosine increase, cosine decrease, Beta-shaped schedules, and inverted-Beta schedules.
All heuristic baselines are tuned to their best-performing configurations for each task.

These baselines represent standard heuristic approaches to guidance scheduling and serve as
natural points of comparison for assessing the benefits of adaptive, policy-learned guidance.

\begin{figure*}[t]
\centering
\includegraphics[width=\linewidth]{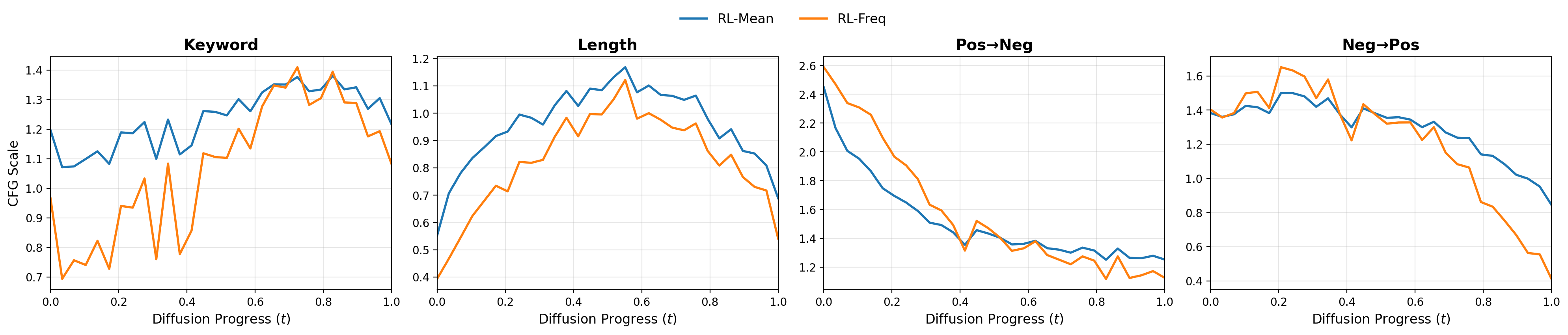}
\caption{Mean guidance trajectories learned by the RL policy across diffusion progress for different tasks under 60-step sampling.}
\label{fig:trajectory_60}
\end{figure*}

\begin{table*}[t]
\centering
\label{tab:main_results_60}

\resizebox{\textwidth}{!}{
\begin{tabular}{lccccccccccc}
\toprule
\multirow{2}{*}{Method}
& \multicolumn{2}{c}{Keyword Generation}
& \multicolumn{3}{c}{Length Control}
& \multicolumn{3}{c}{Sentiment Transfer(Pos2Neg)}
& \multicolumn{3}{c}{Sentiment Transfer(Neg2Pos)}\\
\cmidrule(lr){2-3}
\cmidrule(lr){4-6}
\cmidrule(lr){7-9}
\cmidrule(lr){10-12}
& Ctrl(\%) $\uparrow$ & PPL $\downarrow$
& Acc(\%) $\uparrow$ & Content(\%) $\uparrow$ & PPL $\downarrow$
& Acc(\%) $\uparrow$ & Content(\%) $\uparrow$ & PPL $\downarrow$
& Acc(\%) $\uparrow$ & Content(\%) $\uparrow$ & PPL $\downarrow$\\
\midrule
Fixed CFG 
& 71.4 & 61.3
& 76.0 & 90.4 & 301.9 
& 99.0 & 66.8 & 184.7
& 32.8 & 68.7 & 137.2 \\
Linear Increase
& 73.8 & 62.0
& 85.4 & 89.8 & 374.8 
& 98.6 & \textbf{74.6} & 345.9
& 38.0 & 71.9 & 301.3\\
Linear Decrease
& 66.2 & 89.2
& 65.6 & 86.8 & 396.5 
& 98.4 & 59.9 & 164.9
& 27.0 & 62.9 & 98.7\\
Cosine Increase
& 70.6 & 66.7
& 87.2 & 90.0 & 378.9
& 98.2 & 74.3 & 339.4
& 38.2 & 72.1 & 326.2\\
Cosine Decrease
& 68.0 & 88.3
& 63.2 & 86.1 & 398.9 
& 98.6 & 58.7 & 150.8
& 26.0 & 62.1 & \textbf{95.4}\\
Beta
& 73.8 & 65.0
& 78.0 & 90.0 & 294.7
& 97.6 & 71.7 & 199.3
& \textbf{38.6} & 72.7 &163.1 \\
Inverted Beta
& 68.8 & 76.3
& 75.0 & 86.5 & 542.9
& 97.8 & 60.6 & 241.0
& 29.8 & 63.5 & 141.2\\
\midrule
\textbf{RL-Mean(ours)}
& \textbf{74.6} & 56.2
& \textbf{92.8} & \textbf{91.8} & \textbf{205.6}
& 99.4 & 67.2 & \textbf{145.2}
& 40.2 & 75.9 & 110.1 \\
\textbf{RL-Freq(ours)}
& 74.2 & \textbf{54.6}
& 92.2 & 91.7 & 211.0
& \textbf{99.6} & 63.2 & 150.7
& \textbf{40.6} & \textbf{76.1} & 106.6 \\
\bottomrule
\end{tabular}
}
\caption{Main results on three controlled text generation tasks with 60 diffusion sampling steps. Fixed CFG on keywords, Length, Pos2Neg and Neg2Pos are 1.5, 1, 1.5, 1.5 respective to achieve the best results over all fixed CFG scales. Other heuristic CFG scales ranges from 0 to 3.}
\end{table*}

% ================================================================
  \subsection{Main Results}

  Table~\ref{tab:main_results_60} presents the full comparison across all four tasks with 60 diffusion steps.
  We highlight three key findings.

  \paragraph{RL-learned schedules consistently outperform fixed and heuristic baselines.}
  Across all four tasks, both \textsc{RL-Mean} and \textsc{RL-Freq} either achieve the best performance or remain competitive with the best heuristic on every metric.
  On \textbf{keyword generation}, \textsc{RL-Mean} attains the highest keyword coverage of 74.6\%, a 3.2 percentage point (pp) improvement over the fixed CFG baseline (71.4\%), while simultaneously reducing perplexity from 61.3 to 56.2.
  \textsc{RL-Freq} further improves fluency to a perplexity of 54.6 with comparable coverage (74.2\%).
  On \textbf{length control}, the gains are most pronounced: \textsc{RL-Mean} achieves 92.8\% accuracy, a 16.8\,pp absolute improvement over fixed CFG (76.0\%), while also delivering the best content preservation (91.8\%) and the lowest perplexity (205.6)---a Pareto improvement across all three metrics.
  Even the best-performing heuristic (cosine increase, 87.2\% accuracy) falls 5.6\,pp short and incurs nearly double the perplexity (378.9 vs.\ 205.6).
  For \textbf{pos$\to$neg sentiment transfer}, where all methods already achieve near-saturated accuracy ($\geq$97.6\%), \textsc{RL-Mean} reduces perplexity to 145.2 (vs.\ 184.7 for fixed CFG) while maintaining 99.4\% accuracy, and \textsc{RL-Freq} reaches the highest accuracy of 99.6\%.
  On the more challenging \textbf{neg$\to$pos} direction, \textsc{RL-Freq} achieves 40.6\% accuracy and 76.1\% content similarity---improvements of 7.8\,pp and 7.4\,pp over fixed CFG (32.8\% and 68.7\%), respectively---while also lowering perplexity from 137.2 to 106.6.

  \paragraph{Heuristic schedules exhibit a controllability--fluency trade-off that RL resolves.}
  Among the heuristic baselines, a clear and consistent pattern emerges: increasing schedules (linear increase, cosine increase) improve task controllability but severely degrade fluency, while decreasing schedules exhibit the opposite behavior.
  For instance, on neg$\to$pos sentiment, cosine increase raises accuracy to 38.2\% but inflates perplexity to 326.2; conversely, cosine decrease achieves the lowest perplexity (95.4) but suffers the worst accuracy (26.0\%), a 14.6\,pp drop from \textsc{RL-Freq}.
  On length control, cosine increase reaches 87.2\% accuracy but at a perplexity of 378.9, while linear decrease drops accuracy to 65.6\% with a perplexity of 396.5.
  The beta distribution schedule, which peaks in the middle, generally falls between these two extremes but does not consistently excel on any task.
  In contrast, the RL-learned schedules break this trade-off: they simultaneously improve both controllability and fluency by learning when to apply strong versus weak guidance throughout the generation process.

  \paragraph{Learned trajectories reveal task-dependent guidance strategies.}
  Figure~\ref{fig:trajectory_60} visualizes the CFG scale trajectories produced by \textsc{RL-Mean} and \textsc{RL-Freq} across the four tasks.
  On \textbf{keyword generation} and \textbf{length control}, the learned schedules exhibit a \emph{hump-shaped} pattern: guidance increases during the early-to-middle
  diffusion steps ($t \approx 0.1$--$0.5$), where the overall token structure is being established, then gradually decreases toward the end of generation.
  This suggests that strong guidance is most beneficial during the coarse structure-forming phase, while lighter guidance during the final refinement steps helps preserve
  fluency.
  In contrast, for \textbf{pos$\to$neg sentiment transfer}, the policy learns a \emph{monotonically decreasing} schedule, starting with high CFG scales ($\sim$2.5) that
  rapidly decline.
  This indicates that establishing the target sentiment polarity early is critical, after which the model can focus on generating coherent and fluent content with reduced
  guidance.
  The \textbf{neg$\to$pos} direction follows a similar but more moderate decreasing trend, consistent with this task being inherently harder (reflected by the lower absolute
  accuracy across all methods).
  Notably, \textsc{RL-Freq} consistently uses lower guidance scales than \textsc{RL-Mean}, particularly in later steps, which explains its tendency toward better fluency
  (lower perplexity) on several tasks.

\section{Conclusion}                                              
  We have presented a reinforcement learning framework for learning dynamic classifier-free guidance schedules in discrete diffusion language models. By formulating guidance selection as a sequential decision-making problem with sparse terminal rewards, our approach discovers task-specific guidance trajectories that adapt to the evolving generation state---without modifying the pretrained model.Experiments on keyword-conditioned generation, length-controlled rewriting, and sentiment style transfer show that the learned schedules consistently outperform fixed and heuristic baselines, achieving simultaneous improvements in controllability and fluency that static schedules cannot.
  Analysis of the learned trajectories reveals distinct guidance
  patterns across tasks: hump-shaped profiles for structural
  constraints (keywords, length) and monotonically decreasing
  profiles for stylistic control (sentiment), suggesting that
  optimal guidance allocation is fundamentally task-dependent.

  \paragraph{Limitations.}
  While adaptive guidance scheduling significantly improves
  the controllability--quality trade-off within the discrete
  diffusion paradigm, a gap remains compared to autoregressive
  large language models, which benefit from stronger language
  modeling priors and more mature decoding strategies.
  The guidance mechanism can mitigate but not fully compensate
  for the inherent limitations of current diffusion language
  models, such as token repetition and fluency degradation
  under strong conditioning.

\bibliography{iclr2026_conference}
\bibliographystyle{iclr2026_conference}

\appendix
%===================================================
\section{Experimental Setup}

\subsection{Base Model}
We use LLaDA-8B-Instruct (GSAI-ML/LLaDA-8B-Instruct), an 8-billion-parameter masked diffusion language model. The model is loaded in bfloat16 precision and kept frozen (eval
   mode) throughout training — only the lightweight RL policy network is trained. LLaDA generates text via an iterative denoising process: the generation region is initialized
   entirely with [MASK] tokens, and at each diffusion step, the model predicts logits over the vocabulary for all masked positions. Tokens with the highest prediction
  confidence are unmasked, and this process repeats until all tokens are revealed. We adopt the semi-autoregressive block-based generation scheme, where the generation length
  is divided into blocks that are filled sequentially. Classifier-free guidance (CFG) is applied at each step using the formulation $\text{logits} = \text{logits}_{uc} +
  (\gamma + 1)(\text{logits}c - \text{logits}{uc})$, where $\gamma$ is the CFG scale selected by the RL policy. Throughout all experiments, we use deterministic decoding
  (Gumbel temperature = 0) and the low-confidence remasking strategy.

\subsection{RL Framework}
  We formulate the problem of learning a dynamic CFG schedule as a Markov Decision Process (MDP) and solve it with Proximal Policy Optimization (PPO). At each decision point
  during the diffusion process, the policy observes the current generation state and selects a CFG scale to apply. To reduce the length of the decision horizon while
  preserving diffusion quality, we employ action repeat: the same CFG scale is applied to multiple consecutive diffusion sub-steps per policy decision. All four tasks use 30
  effective decision points.

  The policy and value networks are implemented as separate MLPs with layer normalization. The actor network consists of two hidden layers (128 units each) with LayerNorm and
  ReLU activations, followed by an output head. The critic network shares the same architecture but outputs a scalar state value. Weights are initialized with orthogonal
  initialization (gain = 0.01 for the actor, gain = 1.0 for the critic) to encourage initial exploration.

  We use the TorchRL framework with TensorDict-based data flow. Rollouts are collected on-policy, advantages are computed via Generalized Advantage Estimation (GAE), and the
  policy is updated with clipped surrogate objectives. All tasks use sparse reward, meaning the reward signal is provided only at the end of each episode (i.e., after the full
   diffusion trajectory completes).

\subsection{Tasks and Datasets}
For the keyword-constrained generation and length control tasks, we construct datasets from WikiText-103 ~\citep{merity2016pointer} by extracting sentences and using spaCy ~\citep{honnibal2017spacy}
   for keyword extraction. For sentiment style transfer, we use the Yelp sentiment dataset ~\citep{shen2017style}, a standard benchmark in text style
  transfer research containing non-parallel positive and negative reviews.
  
  We evaluate on three controlled generation tasks spanning four experimental configurations:

  Keyword-Constrained Generation. Given a set of 10 keywords, the model must generate a fluent sentence containing all keywords. The prompt is formatted as: "Generate a fluent
   and coherent sentence that contains all of the following keywords: [keywords]." The dataset contains 3,000 training samples and 500 evaluation samples, each consisting of a
   reference sentence (20–50 words, mean = 34.5) paired with 10 extracted keywords. All 3,000 training samples are used during training.

  Length-Controlled Sentence Compression. Given a sentence, the model must compress it to 40\%–80\% of its original word count while preserving meaning. The prompt is formatted
  as: "Compress to 40\%-80\% length. Output only the compressed sentence, no explanation. Input: [sentence] Output:" The dataset contains 5,000 training samples and 500
  evaluation samples, each with a sentence and its word count. We use 3,000 training samples.

  Sentiment Style Transfer (neg$\rightarrow$pos and pos$\rightarrow$neg). Given a sentence in the source sentiment, the model must rewrite it in the target sentiment while preserving meaning. We train two separate policies for the two transfer directions. The prompt follows the template: "Rewrite the following [source] sentence into a [target]
   sentence while preserving the original meaning." The sentiment dataset contains 50,000 training sentences per polarity (from which we sample 3,000) and 500 development
  sentences per polarity for evaluation.

\subsection{State Space}

  1. Step ratio ($t / T$): the fraction of completed diffusion steps, providing a temporal signal.
  
  2. Mask ratio: the fraction of generation tokens still masked, indicating generation progress.
  
  3. Task-specific progress: keyword coverage ratio (fraction of keywords found in current text), compression length ratio (current word count / original word count), or
  sentiment score (binary: whether current text is classified as the target sentiment).
  
  4. Previous CFG scale (normalized): the CFG scale used at the previous decision point, enabling the policy to condition on its own recent actions.
  
  5. Model confidence: the mean softmax probability of predicted tokens at masked positions, reflecting how certain the model is about its current predictions.

  All features are normalized to approximately [0, 1].

\subsection{Action Space}

  For the three tasks, the action space is discrete with 13 choices: {0.0, 0.25, 0.50, \ldots, 3.0}, parameterized by a Categorical distribution over logits. The discrete granularity of 0.25 provides sufficient resolution for CFG control while keeping the action space tractable for policy learning.

\subsection{Reward Design}
  All tasks combine task-specific quality metrics with a fluency measure (GPT-2 perplexity). Perplexity rewards are computed as $r_{ppl} = 1 -
  \text{clip}\left(\frac{\text{PPL} - 1}{\text{PPL}_{max} - 1}, 0, 1\right)$, capped at a task-specific maximum.

  Keyword generation: $R = 0.5 \cdot r_{coverage} + 0.5 \cdot r_{ppl}$, where $r_{coverage}$ is a strict binary indicator (1.0 if all 10 keywords appear as exact word matches,
   0.0 otherwise), and $\text{PPL}_{max} = 120$.

  Length control: $R = 0.45 \cdot r_{length} + 0.10 \cdot r_{content} + 0.45 \cdot r_{ppl}$, where $r_{length}$ is 1.0 if the compressed word count falls within the target
  range and decays linearly outside it, $r_{content}$ is the cosine similarity between Sentence-BERT embeddings of the original and compressed sentences, and $\text{PPL}_{max}
   = 500$.

  Sentiment transfer (neg$\rightarrow$pos): $R = 0.6 \cdot r_{cls} + 0.3 \cdot r_{ppl} + 0.1 \cdot r_{semantic}$, where $r_{cls}$ is the target-class probability from a fine-tuned RoBERTa sentiment
  classifier (accuracy = 0.97), and $\text{PPL}_{max} = 300$.

  Sentiment transfer (pos$\rightarrow$neg): $R = 0.3 \cdot r_{cls} + 0.6 \cdot r_{ppl} + 0.1 \cdot r_{semantic}$ , with the same classifier and $\text{PPL}_{max} = 300$. The higher PPL weight for this
  direction reflects the empirical observation that negative-style generation requires stronger fluency regularization.

\subsection{Heuristic Schedules}
\begin{figure*}[t]
\centering
\includegraphics[width=0.5\linewidth]{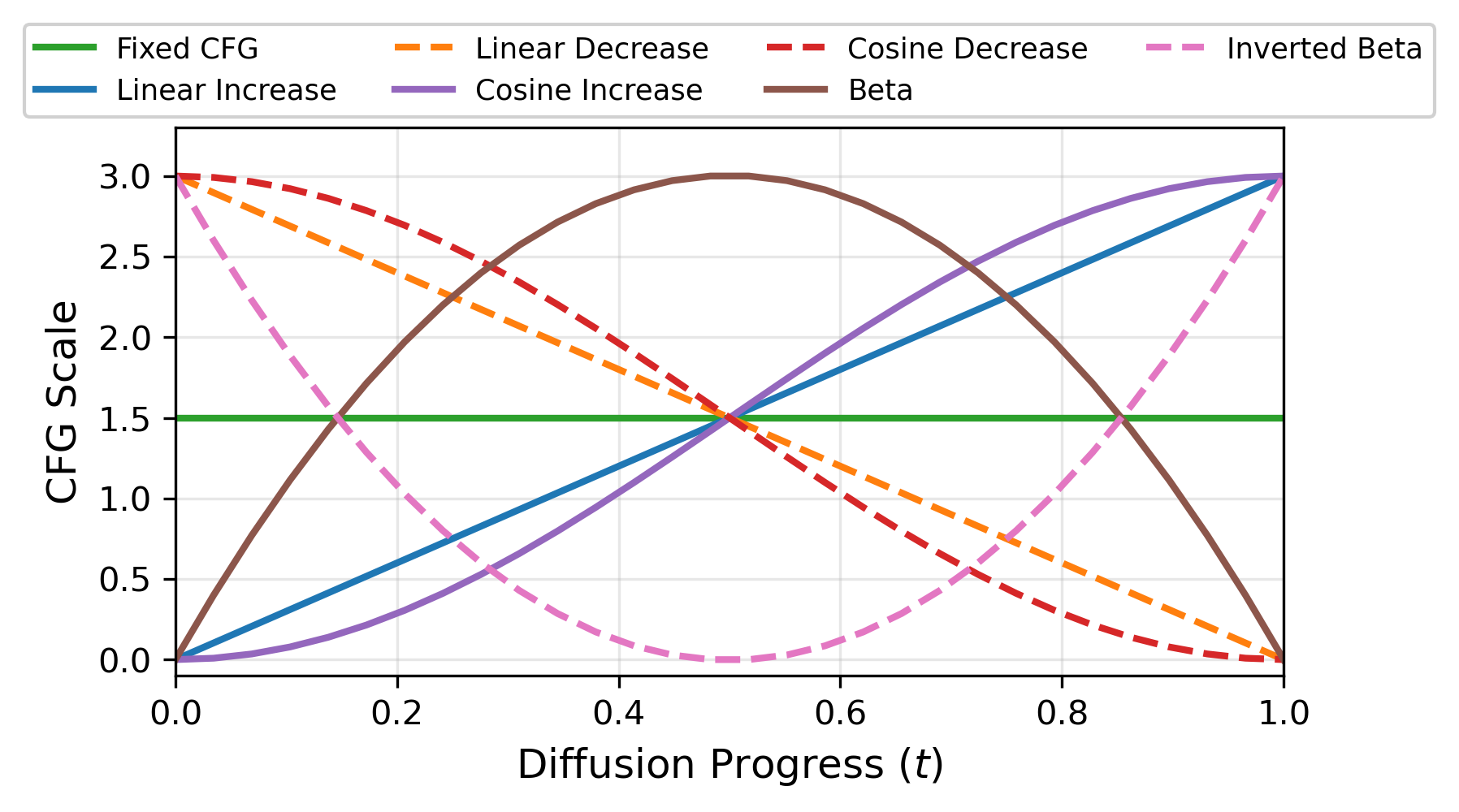}
\caption{Seven heuristic guidance schedules used as baselines.                                                                                                   
  All schedules operate within $[0, \gamma_{\max}]$ where                                                                                                          
  $\gamma_{\max}=3.0$. Fixed CFG uses a constant scale of 1.5.
  Increasing and decreasing variants are defined for linear and
  cosine families. The beta schedule follows a $\mathrm{Beta}(2,2)$
  density, peaking at the midpoint, while inverted beta is its
  complement.}
\label{heuristic_curves}
\end{figure*}

 To contextualize the learned guidance schedules, we compare
  against seven heuristic baselines (Figure~\ref{heuristic_curves}).
  \textbf{Fixed CFG} applies different fixed CFG scale over different tasks
  throughout generation.
  \textbf{Linear} and \textbf{cosine} variants monotonically
  increase or decrease the guidance scale between $0$ and
  $\gamma_{\max}=3.0$, representing the intuition that guidance
  should either strengthen as generation progresses (to reinforce
  constraints on partially formed tokens) or weaken (to avoid
  disrupting already-denoised content).
  The \textbf{beta} schedule, parameterized as a
  $\mathrm{Beta}(2,2)$ density scaled to $[0, \gamma_{\max}]$,
  concentrates guidance in the middle diffusion steps, while
  \textbf{inverted beta} applies strong guidance at both
  endpoints and minimal guidance at the midpoint.
  Together, these baselines span a diverse set of monotonic,
  symmetric, and constant guidance strategies.

%===================================================
\section{Sensitivity Analysis}

\subsection{Effect of Stochastic Temperature}

\begin{figure*}[t]
\centering
\includegraphics[width=\linewidth]{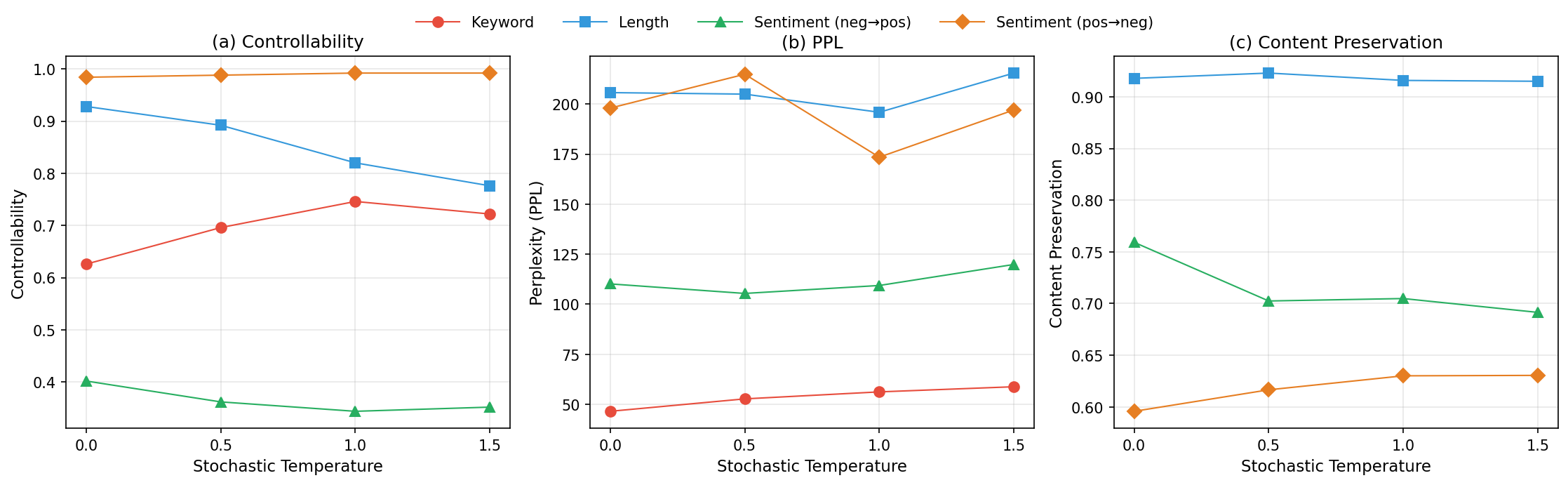}
\caption{Ablation study on policy sampling temperature. We report controllability, fluency (GPT-2 perplexity), and content preservation across tasks.}
\label{fig:temperature_ablation}
\end{figure*}

We vary the sampling temperature of the guidance policy to control the level of stochasticity during trajectory sampling. Figure \ref{fig:temperature_ablation} shows the effect of temperature on controllability, fluency (PPL), and content preservation.

As temperature increases, controllability generally improves due to increased exploration of stronger guidance actions, while fluency and content preservation gradually degrade. Moderate temperatures achieve the best tradeoff, indicating that limited stochastic exploration is beneficial for discovering effective guidance behaviors, whereas excessive randomness leads to unstable or overly aggressive guidance.

\subsection{Effect of Reward Weights}

\begin{figure*}[t]
\centering
\includegraphics[width=0.5\linewidth]{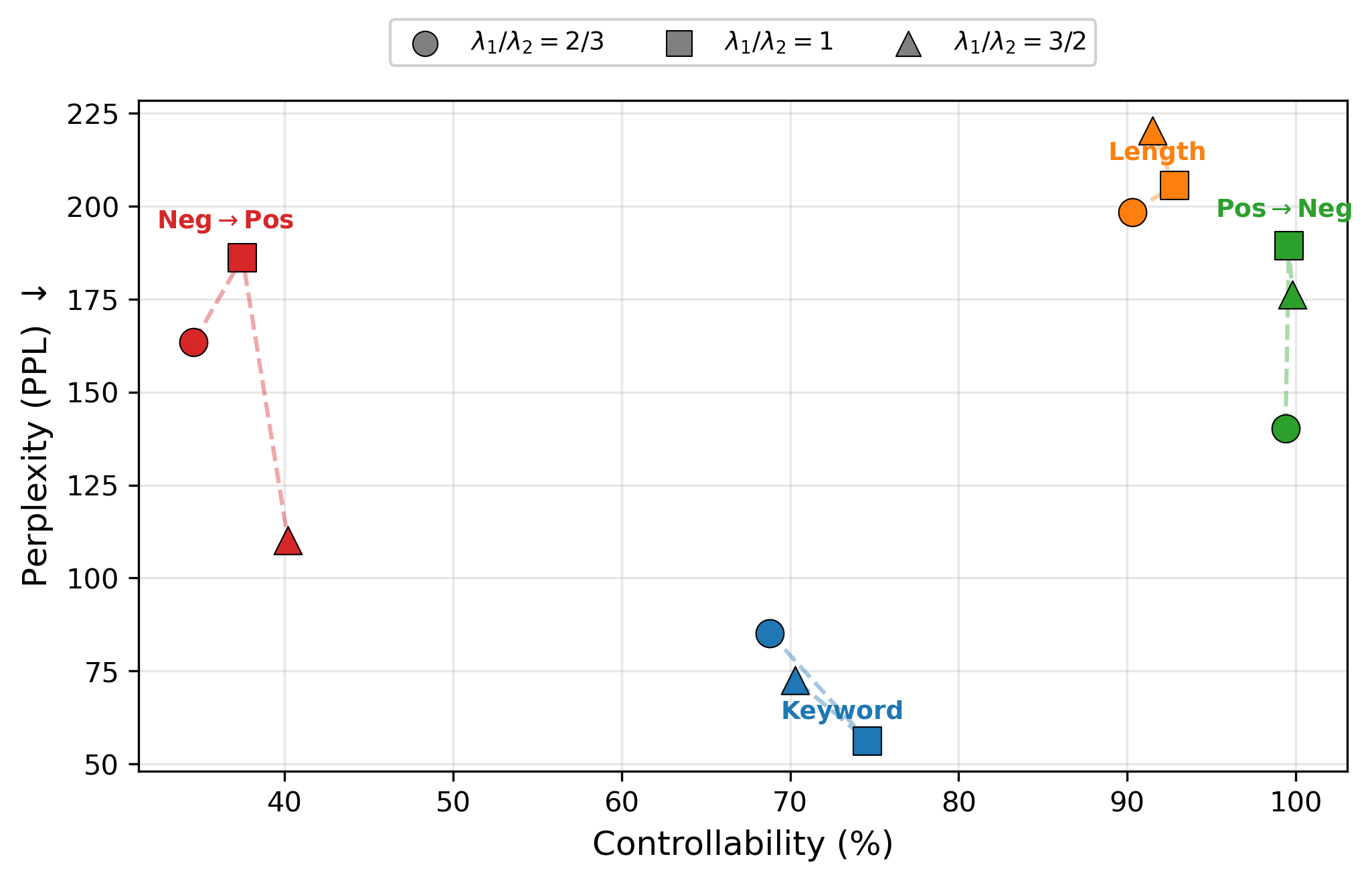}
\caption{Controllability--fluency Pareto front under different                                                                                                   
  reward weight ratios ($\lambda_1/\lambda_2$) across four tasks.                                                                                                  
  Each task traces a trade-off curve: higher $\lambda_1/\lambda_2$                                                                                                 
  improves controllability at the cost of fluency.} 
\label{fig:reward_pareto}
\end{figure*}

  Figure~\ref{fig:reward_pareto} visualizes the
  controllability--fluency trade-off under different reward                                                                                                        
  weight ratios $\lambda_1/\lambda_2 \in \{2/3, 1, 3/2\}$.
  The optimal ratio differs across tasks and reflects the                                                                                                          
  inherent difficulty of each controllability objective.For \textbf{pos$\to$neg} sentiment transfer, where all
  configurations already achieve near-perfect accuracy
  ($\geq$99.4\%), controllability is effectively saturated.
  Accordingly, allocating more weight to fluency
  ($\lambda_1/\lambda_2 = 2/3$) yields the best overall
  result, as the policy can focus its optimization budget
  on reducing perplexity without sacrificing accuracy.
  Conversely, for \textbf{neg$\to$pos}---the hardest task
  with accuracy ranging from 34.6\% to 40.2\%---a higher
  controllability weight ($\lambda_1/\lambda_2 = 3/2$)
  is necessary to push the policy toward stronger guidance
  that improves sentiment transfer accuracy.
  For \textbf{keyword generation} and \textbf{length control},
  a balanced ratio ($\lambda_1/\lambda_2 = 1$) achieves
  the best trade-off, as neither controllability nor fluency
  is trivially solved.

\subsection{Effect of Number of Diffusion Sampling Timesteps}

  We compare adaptive guidance learned under 30-step and 60-step
  diffusion sampling (Tables~\ref{tab:main_results_30}
  and~\ref{tab:main_results_60}). The effect of increasing the
  number of diffusion steps varies across tasks.
  On \textbf{length control} and \textbf{neg$\to$pos sentiment},
  more steps consistently improve all metrics---for instance,
  RL-Mean accuracy on length control increases from 83.8\% to
  92.8\% while perplexity drops from 222.7 to 205.6.
  On \textbf{keyword generation}, additional steps substantially
  improve fluency (RL-Mean PPL: 87.6$\to$56.2) but slightly
  reduce keyword coverage (78.8\%$\to$74.6\%), suggesting a
  trade-off between controllability and fluency as the sampling
  horizon lengthens.
  On \textbf{pos$\to$neg sentiment}, accuracy and content
  preservation improve, while perplexity slightly increases
  for the RL methods (RL-Mean: 116.6$\to$145.2), indicating
  that the policy leverages additional steps primarily to
  strengthen style transfer at a modest fluency cost.

\begin{figure*}[t]
\centering
\includegraphics[width=\linewidth]{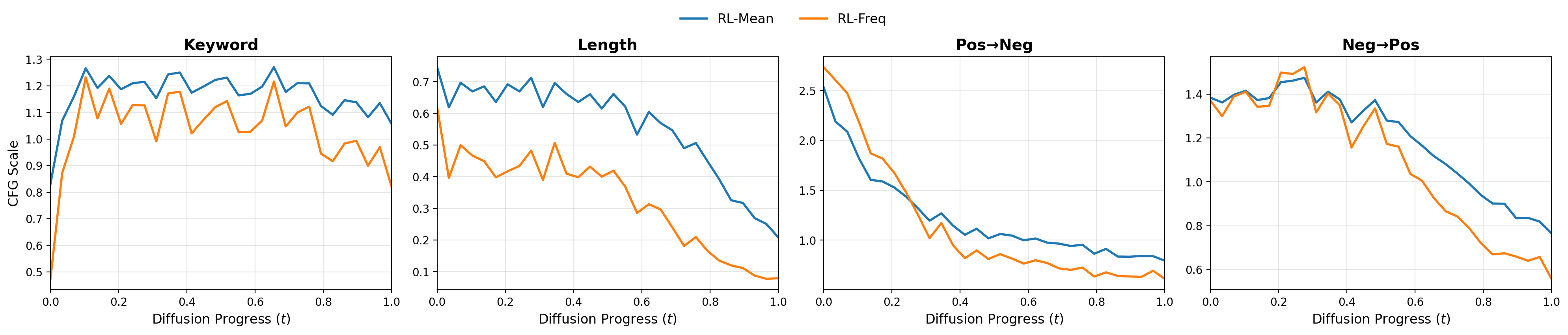}
\caption{Mean guidance trajectories learned by the RL policy across diffusion progress for different tasks under 30-step sampling.}
\label{fig:trajectory_30}
\end{figure*}

  Notably, while the qualitative task-dependent patterns of the
  learned guidance trajectories are preserved across sampling
  horizons (Figures~\ref{fig:trajectory_30} and~\ref{fig:trajectory_60}),
  the specific scale values and trajectory shapes differ,
  indicating that the policy adapts its guidance strategy to
  the available number of diffusion steps rather than simply
  rescaling a fixed schedule.

%===================================================
\section{Case Studies}
 To provide qualitative insight into how different guidance strategies affect generation quality, we present representative outputs from all methods on each task (Table \ref{keywords_30} - Table \ref{neg2pos_60}).

\begin{table*}[t]
\centering
\label{tab:main_results_30}

\resizebox{\textwidth}{!}{
\begin{tabular}{lccccccccccc}
\toprule
\multirow{2}{*}{Method}
& \multicolumn{2}{c}{Keyword Generation}
& \multicolumn{3}{c}{Length Control}
& \multicolumn{3}{c}{Sentiment Transfer(Pos2Neg)}
& \multicolumn{3}{c}{Sentiment Transfer(Neg2Pos)}\\
\cmidrule(lr){2-3}
\cmidrule(lr){4-6}
\cmidrule(lr){7-9}
\cmidrule(lr){10-12}
& Ctrl(\%) $\uparrow$ & PPL $\downarrow$
& Acc(\%) $\uparrow$ & Content(\%) $\uparrow$ & PPL $\downarrow$
& Acc(\%) $\uparrow$ & Content(\%) $\uparrow$ & PPL $\downarrow$
& Acc(\%) $\uparrow$ & Content(\%) $\uparrow$ & PPL $\downarrow$\\
\midrule
Fixed CFG 
& 74.2 & 107.6
& 70.8 & 89.4 &  395.4
& 98.4 & 63.9 & 196.2
& 31.2 & 66.1 & 162.7 \\
Linear Increase
& 76.8 & 150.1
& 83.0 & 88.9  &  522.5
& 98.4 & 73.0 & 412.7
& \textbf{41.8} & 72.5 & 378.9 \\
Linear Decrease
& 72.6 & 111.9
& 54.6 & 85.6 &  393.2
& \textbf{99.6} & 56.6 & 122.3
& 19.6 & 60.0 & \textbf{96.4} \\
Cosine Increase
& 74.8 & 169.3
& 84.2 & 88.3 & 551.4  
& 97.8 & \textbf{73.7} & 422.0
& \textbf{41.8} & 71.6 & 391.6 \\
Cosine Decrease
& 70.6 & 110.6
& 53.4 & 84.7 & 406.0 
& \textbf{99.6} & 56.6 & 127.7
& 22.8 & 59.9 & 96.7 \\
Beta
& \textbf{78.8} & 157.2
& 68.0 & 89.0 & 406.4
& 97.8 & 70.3 & 249.7
& 39.2 & 70.4 & 212.6 \\
Inverted Beta
& 75.2 & 86.9
& 67.0 & 86.0 & 550.5
& 99.0 & 57.7 &179.6
& 26.2 & 60.4 &130.7 \\
\midrule
\textbf{RL-Mean(ours)}
& \textbf{78.8} & 87.6
& 83.8 & \textbf{91.5} & \textbf{222.7}
& 98.6 & 59.5 & \textbf{116.6}
& 40.1 & \textbf{72.6} &  126.7\\
\textbf{RL-Freq(ours)}
& 77.0 & \textbf{81.0}
& \textbf{90.0} & 90.1 & 228.8 
& \textbf{99.6}& 62.0 & 122.6
& 39.4 & 61.2 & 121.1 \\
\bottomrule
\end{tabular}
}
\caption{Main results on three controlled text generation tasks with 30 diffusion sampling steps.}
\end{table*}

\begin{table*}[t]
\centering
\footnotesize
\setlength{\tabcolsep}{6pt}
\begin{tabular}{l p{11.2cm}}
\toprule

% ===== Task 1 =====
\multicolumn{2}{p{12.8cm}}{
\textbf{Keyword Generation Prompt.}
Use the following 10 keywords to generate one fluent sentence:
\texttt{entered, tons, long, flooding, meaning, hole, wounded, ship, water, men}.
} \\
\midrule
\textbf{Method} & \textbf{Generated Text} \\
\midrule
Fixed CFG & As the ship entered entered into the long ocean, carrying tons tons long tons of valuable cargo, a sudden storm caused flooding flooding flooding meaning meaning that the ship was at risk of water water water water due to a hidden hole hole that threatened to wounded men men men among the ship ship's water water wounded men. \\
Linear Increase & As the old ship entered the treacherous waters, it encountered tons upon tons of debris, causing a long and relentless flooding that meant a significant hole where it had once sturdy, ultimately leaving leaving the wounded and injured ship clinging to a precarious existence, while its brave men desperately fought for survival amidst the encroaching water. \\
Linear Decrease & As the ship entered entered into the long long long ocean carrying tons tons long tons tons, it encountered a long flooding long flooding event, meaning meaning that the ship ship had entered entered a hole hole long hole hole where wounded men wounded men had been surrounded by water water long water water water wounded men wounded men. \\
Cosine Increase & As the long ship entered the treacherous waters, it massive hull began to flood with tons of water, meaning that it had encountered a catastrophic hole that threatened to engulf the entire vessel, leaving its wounded men to scramble desperately for safety as they fought desperately to keep the ship afloat amidst the relentless, encroaching water. \\
Cosine Decrease & As the ship entered entered the long tons tons tons long tons tons of water, it began flooding flooding revealing flooding flooding revealing flooding flooding revealing revealing revealing revealing flooding revealing revealing revealing a hidden hole meaning hole filled hole filled men wounded men men men men men men men wounded men men who were desperately in need of help. \\
Beta & As the old ship entered entered the treacherous coastline, it encountered tons tons long long long term flooding flooding, meaning meaning that once a small hole gaping hole in the hole was now spilling ship tons water water and threatening ship men wounded men who had been bravely fighting but were injured during the harrowing ordeal. \\
Inverted Beta & As the ship entered entered into the long long coastline, it was flooded with tons tons of water, causing a large hole in the hull, which meant that the ship was at risk of being further wounded by water rising water rising water, while the crew worked tirelessly to repair the hole to save the wounded men. \\
\textbf{RL-Mean (ours)} & As the old ship entered the treacherous waters, it encountered tons upon tons of water that threatened to create a long and dangerous hole in the hole, which meant that the ship was flooding at an alarming rate, posing a significant danger to the wounded men who were desperately trying to stay afloat amidst the rapidly rising water. \\
\textbf{RL-Freq (ours)} & As the old ship entered the treacherous waters after a long voyage, tons of water seeped in, causing severe flooding and a gaping hole, which meant that the wounded men on board had no choice but to fight to save themselves and the ship, as the rising water threatened threatened to engulf the entire vessel. \\

\bottomrule
\end{tabular}

\caption{
Qualitative results for the keyword-conditioned generation task with \textbf{30 diffusion steps}.
All methods are evaluated under the same keyword prompt.
Adaptive guidance (RL-Mean and RL-Freq) produces more coherent sentences with fewer repetitions
compared to fixed and heuristic guidance schedules.
}
\label{keywords_30}
\end{table*}

\begin{table*}[t]
\centering
\footnotesize
\setlength{\tabcolsep}{6pt}
\begin{tabular}{l p{11.2cm}}
\toprule

% ===== Task 1 =====
\multicolumn{2}{p{12.8cm}}{
\textbf{Keyword Generation Prompt.}
Use the following 10 keywords to generate one fluent sentence:
\texttt{entered, tons, long, flooding, meaning, hole, wounded, ship, water, men}.
} \\
\midrule
\textbf{Method} & \textbf{Generated Text} \\
\midrule
Fixed CFG & As the old ship entered the stormy waters, carrying tons of water and causing severe flooding due to a long hole in its hull, it meant that the wounded men on board desperately needed to find a way to escape the rising water before the ship could completely sink, meaning their very lives hung in the balance. \\
Linear Increase & As the long ship entered the treacherous waters, it encountered tons of water that threatened to cause catastrophic flooding, meaning that a significant hole had developed in its hull, severely endangering the wounded men who were desperately navigating the ship's survival amidst the relentless influx of water threatening to engulf their beloved vessel. \\
Linear Decrease & As the ship entered entered tons of water during its long and harrowing journey, a sudden flooding occurred, meaning that a gaping hole had developed in its fragile hull, forcing the wounded men to scramble for safety amidst the rising water engulfing the ship, as they desperately tried to navigate through the treacherous waters. \\
Cosine Increase & As the old ship entered the treacherous waters, a massive hole in its hull began to frantically swallowing tons of water, causing a long and alarming flooding that meant imminent danger for the men on board, including several wounded men as they desperately fought to stabilize the stricken vessel amidst the relentless, encroaching water. \\
Cosine Decrease & As the ship entered entered the harbor, carrying tons of cargo, its long hull was suddenly compromised by a flooding breach, meaning that a gaping hole had developed, allowing massive amounts of water to enter, threatening the safety of the men on board, including wounded individuals who desperately needed to escape the treacherous situation. \\
Beta & As the old ship entered the treacherous stormy sea, tons of water began to pour in due to a disarmingly long hole, which meant that the-mean-meaning crew members, including several wounded men, were desperately trying to navigate the treacherous vessel while also fighting for their lives amidst the perilous flooding. \\
Inverted Beta & As the ship entered entered the harbor, it was carrying tons of cargo and long passengers, but unfortunately, a large hole had developed in its hull, causing severe flooding, which meant that the wounded men on the ship were now in grave danger as they desperately fought to escape and stay afloat amidst the rising water. \\
\textbf{RL-Mean (ours)} & As the wounded ship entered the treacherous waters, a long hole began to widen, causing tons of water to flood in and threatening the lives of the men on board, meaning that the vessel was now in dire need of immediate attention to prevent further damage and ensure the safety of its crew amidst the perilous situation. \\
\textbf{RL-Freq (ours)} & As the long ship entered the treacherous waters, it encountered tons of water pouring in, causing severe flooding that ultimately meant the discovery of a hole in the ship's hull, further endangering the wounded men who now desperately needed to find a way to escape the rising water that threatened to engulf the entire vessel. \\

\bottomrule
\end{tabular}

\caption{
Qualitative results for the keyword-conditioned generation task with \textbf{60 diffusion steps}.
Compared to 30-step sampling, longer diffusion trajectories amplify the differences between guidance strategies.
Learned adaptive guidance yields more fluent and semantically consistent outputs.
}
\label{keywords_60}
\end{table*}

\begin{table*}[t]
\centering
\footnotesize
\setlength{\tabcolsep}{6pt}
\begin{tabular}{l p{11.2cm}}
\toprule

% ===== Task 2 =====
\multicolumn{2}{p{12.8cm}}{
\textbf{Length Control Prompt.}
Rewrite the input sentence so that the output length is within
\textbf{40\%--80\%} of the original length.
\textit{Input:} ``Pennsylvania Representative Joseph Ingersoll , a Whig , spoke against the bill , noting that Patterson opposed the new denominations , and that the idea had been repeatedly turned down , whenever considered.''}\\
\midrule
\textbf{Method} & \textbf{Generated Text} \\
\midrule
Fixed CFG & Pennsylvania Representative Joseph Ingersoll spoke against the bill, noting Patterson Patterson Patterson Patterson Patterson opposed the new denominations, and the idea had been repeatedly turned down whenever considered. \\
Linear Increase & 'unsylvania Whig Ingersoll criticized the bill, citing Patterson's opposition to new denominations and repeated idea idea whenever considered.' \\
Linear Decrease & Representative Ingersgersoll opposed Whig denominations bill, citing Patterson Patterson's opposition and noting the idea had been repeatedly rejected. \\
Cosine Increase & Wh Pennsylvania Whig Ingersoll spoke against bill, citing Patterson's opposition to new denominations and noting the idea idea'd repeatedly turned down whenever considered. \\
Cosine Decrease & Representative Ingersgersoll opposed Whig denominations bill, citing Patterson Patterson's opposition, noting the idea repeatedly repeatedly rejected. \\
Beta & Representative Ingersoll Pennsylvania Whig bill bill. Patterson opposed Patterson new denominations. Idea turned down repeatedly whenever repeatedly whenever considered. \\
Inverted Beta & Representative Ingersgersoll Joseph Whig Pennsylvania spoke bill, noting Patterson Patterson opposed new denominations. Idea had idea repeatedly turned down repeatedly whenever considered. \\
\textbf{RL-Mean (ours)} & Representative Joseph Ingersoll, a Whig, spoke against the bill bill, citing Patterson's opposition to new denominations and repeated rejections. \\
\textbf{RL-Freq (ours)} & Whig Representative Joseph Ingersoll opposed the bill, citing Patterson's stance and repeated rejections. \\
\bottomrule
\end{tabular}

\caption{
Qualitative results for the length-controlled rewriting task with \textbf{30 diffusion steps}.
All methods are prompted to rewrite the input sentence within the target length range (40\%--80\% of the original).
Adaptive guidance better preserves semantic content while satisfying length constraints.}
\label{tab:qualitative_all_tasks_30}
\end{table*}

\begin{table*}[t]
\centering
\footnotesize
\setlength{\tabcolsep}{6pt}
\begin{tabular}{l p{11.2cm}}
\toprule

% ===== Task 2 =====
\multicolumn{2}{p{12.8cm}}{
\textbf{Length Control Prompt.}
Rewrite the input sentence so that the output length is within
\textbf{40\%--80\%} of the original length.
\textit{Input:} ``Pennsylvania Representative Joseph Ingersoll , a Whig , spoke against the bill , noting that Patterson opposed the new denominations , and that the idea had been repeatedly turned down , whenever considered.''}\\
\midrule
\textbf{Method} & \textbf{Generated Text} \\
\midrule
Fixed CFG & Whig Rep Joseph Ingersoll opposed Pennsylvania bill, citing Patterson Patterson's opposition to new denominations and noting the idea had been repeatedly rejected. \\
Linear Increase & 'unsylvania Whig Ingersoll spoke against bill, referencing Patterson's opposition to new denominations and noting idea turned down repeatedly whenever considered.' \\
Linear Decrease & Representative Ingersoll opposed Patterson bill, stating Patterson disapproved of new denominations, idea repeatedly rejected. \\
Cosine Increase & 'unsylvania Whig Ingersoll spoke against bill, referencing Patterson's opposition to new denominations and noting idea turned down repeatedly whenever considered.' \\
Cosine Decrease & Representative Ingersoll opposed Patterson bill, stating Patterson disapproved of new denominations, idea repeatedly rejected. \\
Beta & Representative Ingersoll (Whig) opposed Pennsylvania bill, citing Patterson's opposition to new denominations and noting the idea was repeatedly rejected. \\
Inverted Beta & Representative Ingersoll opposed Pennsylvania bill, citing Patterson's stance to new denominations and noting the idea was repeatedly rejected.\\
\textbf{RL-Mean (ours)} & Representative Ingersoll spoke against the bill, citing Patterson's opposition to new denominations and noting that the idea had been repeatedly rejected. \\
\textbf{RL-Freq (ours)} & Representative Ingersoll spoke against the bill, citing Patterson's opposition to new denominations and noting that the idea had been repeatedly rejected. \\
\bottomrule
\end{tabular}

\caption{
Qualitative results for the length-controlled rewriting task with \textbf{60 diffusion steps}.
With longer diffusion trajectories, adaptive guidance produces more concise and faithful rewrites,
while heuristic schedules often introduce redundancy or fragmentation.
}
\label{tab:qualitative_all_tasks_30}
\end{table*}

\begin{table*}[t]
\centering
\footnotesize
\setlength{\tabcolsep}{6pt}
\begin{tabular}{l p{11.2cm}}
\toprule

% ===== Task 3 =====
\multicolumn{2}{p{12.8cm}}{
\textbf{Sentiment Transfer (Pos$\rightarrow$Neg).}
Rewrite the sentence to \textbf{negative} sentiment while preserving content.
\textit{Input:} ``the pepper steak is another good one .''}
\\
\midrule
\textbf{Method} & \textbf{Generated Text} \\
\midrule
Fixed CFG & pepper steak is another bad one . \\
Linear Increase & The pepper steak is not another good one . \\
Linear Decrease & pepper steak is another bad one . \\
Cosine Increase & the pepper steak lacks another good one . \\
Cosine Decrease & pepper steak is another bad one . \\
Beta & The pepper steak is not another good one. \\
Inverted Beta & pepper steak: another mediocre one . \\
\textbf{RL-Mean (ours)} & pepper steak is another bad one . \\
\textbf{RL-Freq (ours)} & pepper steak is another bad one . \\
\bottomrule
\end{tabular}

\caption{
Qualitative results for sentiment style transfer from positive to negative with \textbf{30 diffusion steps}.
All methods are evaluated on the same input sentence.
Adaptive guidance achieves correct sentiment transfer with minimal distortion of the original content.
}
\label{tab:qualitative_all_tasks_30}
\end{table*}

\begin{table*}[t]
\centering
\footnotesize
\setlength{\tabcolsep}{6pt}
\begin{tabular}{l p{11.2cm}}
\toprule

% ===== Task 3 =====
\multicolumn{2}{p{12.8cm}}{
\textbf{Sentiment Transfer (Pos$\rightarrow$Neg).}
Rewrite the sentence to \textbf{negative} sentiment while preserving content.
\textit{Input:} ``the pepper steak is another good one .''}
\\
\midrule
\textbf{Method} & \textbf{Generated Text} \\
\midrule
Fixed CFG & pepper steak is another bad one . \\
Linear Increase & the pepper steak is not another good one . \\
Linear Decrease & pepper steak is another bad one . \\
Cosine Increase & the pepper steak isn't another good one . \\
Cosine Decrease & pepper steak is another bad one . \\
Beta & The pepper steak isn't another good one. \\
Inverted Beta & pepper steak: another mediocre one . \\
\textbf{RL-Mean (ours)} & pepper steak is another bad one . \\
\textbf{RL-Freq (ours)} & pepper steak is another bad one . \\
\bottomrule
\end{tabular}

\caption{
Qualitative results for sentiment style transfer from positive to negative with \textbf{60 diffusion steps}.
Compared to shorter trajectories, longer sampling further stabilizes sentiment control
under adaptive guidance.
}

\label{tab:qualitative_all_tasks_30}
\end{table*}

\clearpage

\begin{table*}[t]
\centering
\footnotesize
\setlength{\tabcolsep}{6pt}
\begin{tabular}{l p{11.2cm}}
\toprule

% ===== Task 4 =====
\multicolumn{2}{p{12.8cm}}{
\textbf{Sentiment Transfer (Neg$\rightarrow$Pos).}
Rewrite the sentence to \textbf{positive} sentiment while preserving content.
\textit{Input:} ``never will i be back to this place ''}
\\
\midrule
Fixed CFG & I will revisit revisit this place at some point.AlternativelyAlternatively:Positive: I will definitely be back to this place in the future, although it may not be under the original circumstances.Both rewritten sentences change the negative sentence into a positive sentence while preserving the original meaning. \\
Linear Increase &  I shall never revisit this place.\\
Linear Decrease & Positive: revisit revisit revisit revisit this place .(I chose this rewritten sentence to preserve the original meaning of the negative sentence by using \"reisit\" to convey that the speaker speaker will not return or revisit the place. This replaces the negative word \"never\" with a positive action of revisiting.) \\
Cosine Increase & I shall never revisit this place. \\
Cosine Decrease & Positive: never revisit revisit revisit this place .Or can be rewritten as: I will not return to this place.Or can be rewritten as: I will not return to this place.(Note, I have rewritten the negative sentence into a positive sentence while preserving the original meaning) \\
Beta & I will definitely return revisit this place. \\
Inverted Beta & Positive: never revisit revisit revisit this place .Alternatively, you could also use:Positive: I will not return to this location.OrPositive: I will not be here again.These sentences transform the negative sentence into a positive sentence while preserving its original meaning. \\
\textbf{RL-Mean (ours)} & I will never revisit this place. \\
\textbf{RL-Freq (ours)} & I will never revisit this place. \\
\bottomrule
\end{tabular}

\caption{
Qualitative results for sentiment style transfer from negative to positive with \textbf{30 diffusion steps}.
Heuristic guidance often leads to inconsistent polarity or repetitive artifacts,
whereas adaptive guidance yields clearer sentiment reversal.
}

\label{tab:qualitative_all_tasks_30}
\end{table*}

\begin{table*}[t]
\centering
\footnotesize
\setlength{\tabcolsep}{6pt}
\begin{tabular}{l p{11.2cm}}
\toprule

% ===== Task 4 =====
\multicolumn{2}{p{12.8cm}}{
\textbf{Sentiment Transfer (Neg$\rightarrow$Pos).}
Rewrite the sentence to \textbf{positive} sentiment while preserving content.
\textit{Input:} ``never will i be back to this place ''}
\\
\midrule
Fixed CFG & I will never revisit this place. \\
Linear Increase &  I no longer revisit this place.\\
Linear Decrease & I revisit revisit this place. \\
Cosine Increase & I shall never revisit this place. \\
Cosine Decrease & Positive: I revisit revisit this place. \\
Beta & I will not revisit this place. I reaffirm my intention to never return to this place or Positive: I choose not to return . \\
Inverted Beta & I revisit revisit this place. \\
\textbf{RL-Mean (ours)} & I will never revisit this place. \\
\textbf{RL-Freq (ours)} & Positive: I will not revisit this place.In this rewritten sentence, \"never\" is replaced with \"not revisit\" to preserve the original meaning that the person will not return to the place. The positive sentence \"I will not revisit this place\" conveys the same idea as the original negative sentence. \\
\bottomrule
\end{tabular}

\caption{
Qualitative results for sentiment style transfer from negative to positive with \textbf{60 diffusion steps}.
Adaptive guidance produces more stable and semantically faithful sentiment inversion
under longer diffusion trajectories.
}

\label{neg2pos_60}
\end{table*}

\end{document}

%% file: math_commands.tex
\usepackage{amsmath,amsfonts,bm}

\def\eqref#1{equation~\ref{#1}}
\def\1{\bm{1}}

\DeclareMathAlphabet{\mathsfit}{\encodingdefault}{\sfdefault}{m}{sl}
\SetMathAlphabet{\mathsfit}{bold}{\encodingdefault}{\sfdefault}{bx}{n}